\documentclass[letterpaper, 10 pt, conference]{ieeeconf}  

\IEEEoverridecommandlockouts                              

\usepackage{graphics} 
\usepackage{float}
\usepackage{subfigure}
\usepackage{epsfig} 
\usepackage{amsmath} 
\usepackage{multirow}
\usepackage{stfloats}
\usepackage{color}
\usepackage{enumerate}
\usepackage{diagbox}
\usepackage{threeparttable} 
\usepackage{cite} 
\usepackage{booktabs}
\usepackage{xcolor}
\usepackage{multirow}
\usepackage{blindtext} 
\let\labelindent\relax
\usepackage[inline]{enumitem}
\DeclareMathOperator*{\argmin}{argmin}
\DeclareMathOperator*{\argmax}{argmax}

\title{\LARGE \bf
Monocular Camera Mapping with Pose-Guided Optimization: Enhancing Marking-Level HD Map Accuracy
}

\author{Hongji Liu$^{1}$, Linwei Zheng$^{1}$, Xiaoyang Yan$^{1}$, Zhenhua Xu$^{1}$, Bohuan Xue$^{1}$, Yang Yu$^{2}$ and Ming Liu$^{2,3,4}$
\thanks{$^{1}$Hongji Liu, Linwei Zheng, Xiaoyang Yan, Zhenhua Xu, Bohuan Xue are with The Hong Kong University of Science and Technology, Hong Kong SAR, China. {\tt\small hliucq@connect.ust.hk}}
\thanks{$^{2}$Yang Yu and Ming Liu are with The Hong Kong University of Science and Technology (Guangzhou), Nansha, Guangzhou, 511400, Guangdong, China.}
\thanks{$^{3,4}$Ming Liu is also with The Hong Kong University of Science and Technology, Hong Kong SAR, China, and with HKUST Shenzhen-Hong Kong Collaborative Innovation Research Institute, Futian, Shenzhen. {\tt\small eelium@ust.hk}}
}

\begin{document}
\bstctlcite{IEEEexample:BSTcontrol}
\maketitle
\thispagestyle{empty}
\pagestyle{empty}

\begin{abstract}
Marking-level high-definition maps (HD maps) are of great significance for autonomous vehicles (AVs), especially in large-scale, appearance-changing scenarios where AVs rely on markings for localization and lanes for safe driving. In this paper, we propose a pose-guided optimization framework for automatically building a marking-level HD map with accurate markings positions using a simple sensor setup (one or more monocular cameras). We optimize the position of the marking corners to fit the result of marking segmentation and simultaneously optimize the inverse perspective mapping (IPM) matrix of the corresponding camera to obtain an accurate transformation from the front view image to the bird's-eye view (BEV). In the quantitative evaluation, the built HD map almost attains centimeter-level accuracy. The accuracy of the optimized IPM matrix is similar to that of the manual calibration. The method can also be generalized to build HD maps in a broader sense by increasing the types of recognizable markings. The supplementary materials and videos are available at http://liuhongji.site/V2HDM-Mono/.
\end{abstract}

\section{INTRODUCTION}

\subsection{Motivation}
High-definition maps (HD maps) are of great assistance to many modules of unmanned ground vehicles (UGVs), such as localization, perception, and planning~\cite{seif2016autonomous}, especially in large-scale appearance-changing application scenarios where autonomous vehicles rely on markings for localization and lanes for safe driving~\cite{yu2022accurate}. 
With such demands, the coordinates of markings need to be accurate enough to assist the localization of UGVs. The inverse perspective mapping (IPM) matrix converting camera images to the bird's-eye view (BEV) also needs to be precise enough in order to ensure the calculation accuracy of the distance between the vehicle and lanes. In practical application, the generation of such HD maps is highly dependent on manual measurement with surveying and mapping tools such as Total Station~\cite{yu2022accurate}. However, there are usually a great number of markings in large-scale scenes, which brings a huge manual workload and a time-consuming measurement process. Therefore, it is of interest to the community to build the HD marking map automatically.

\begin{figure}[t] 
    \centering
        \subfigure[Aerial view of the experimental scene]{
        \includegraphics[width=0.45\textwidth]{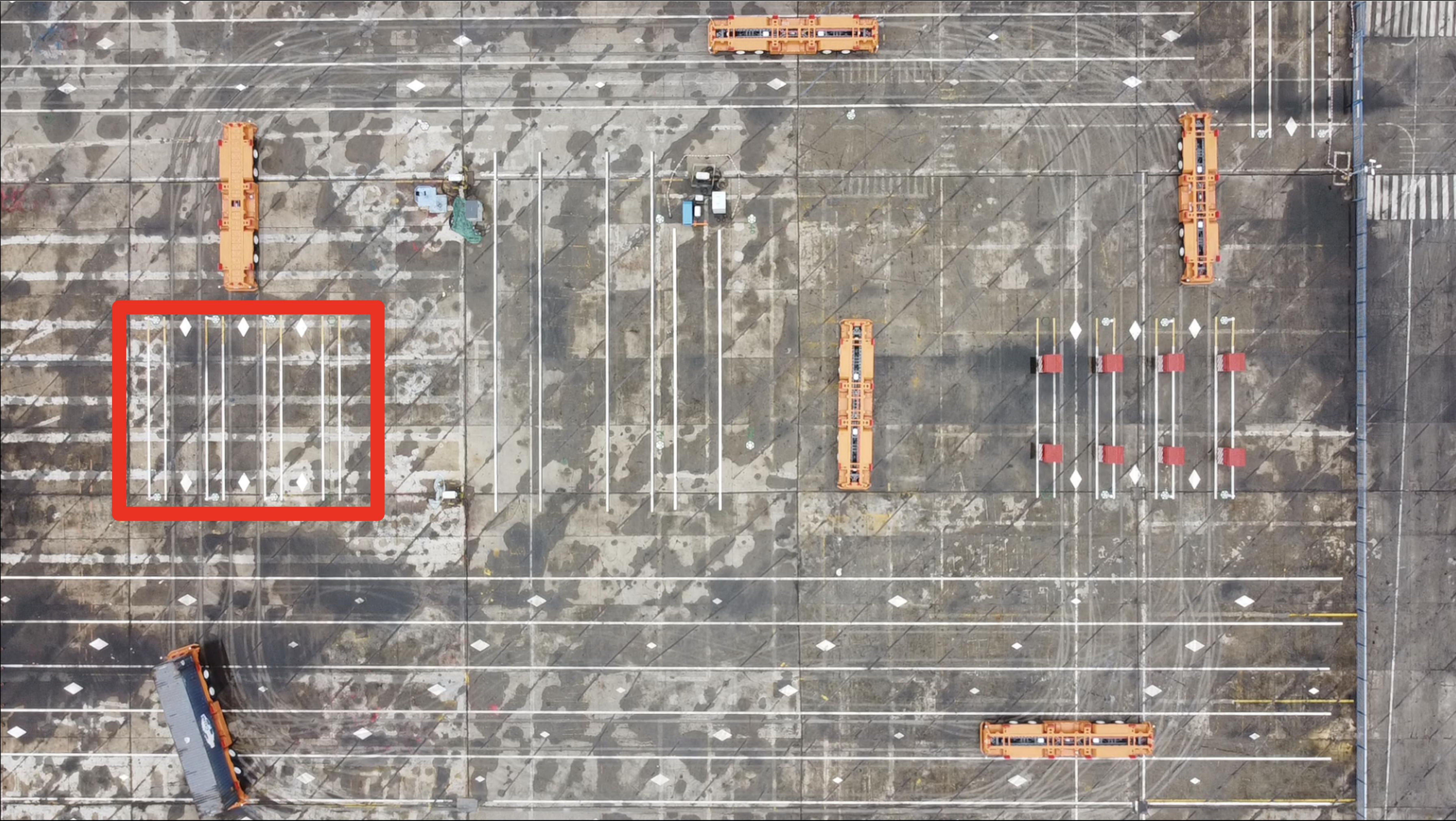}
        \label{fig: main_live}
    }
    \subfigure[Automatically generated partial marking HD map]{
        \includegraphics[width=0.45\textwidth]{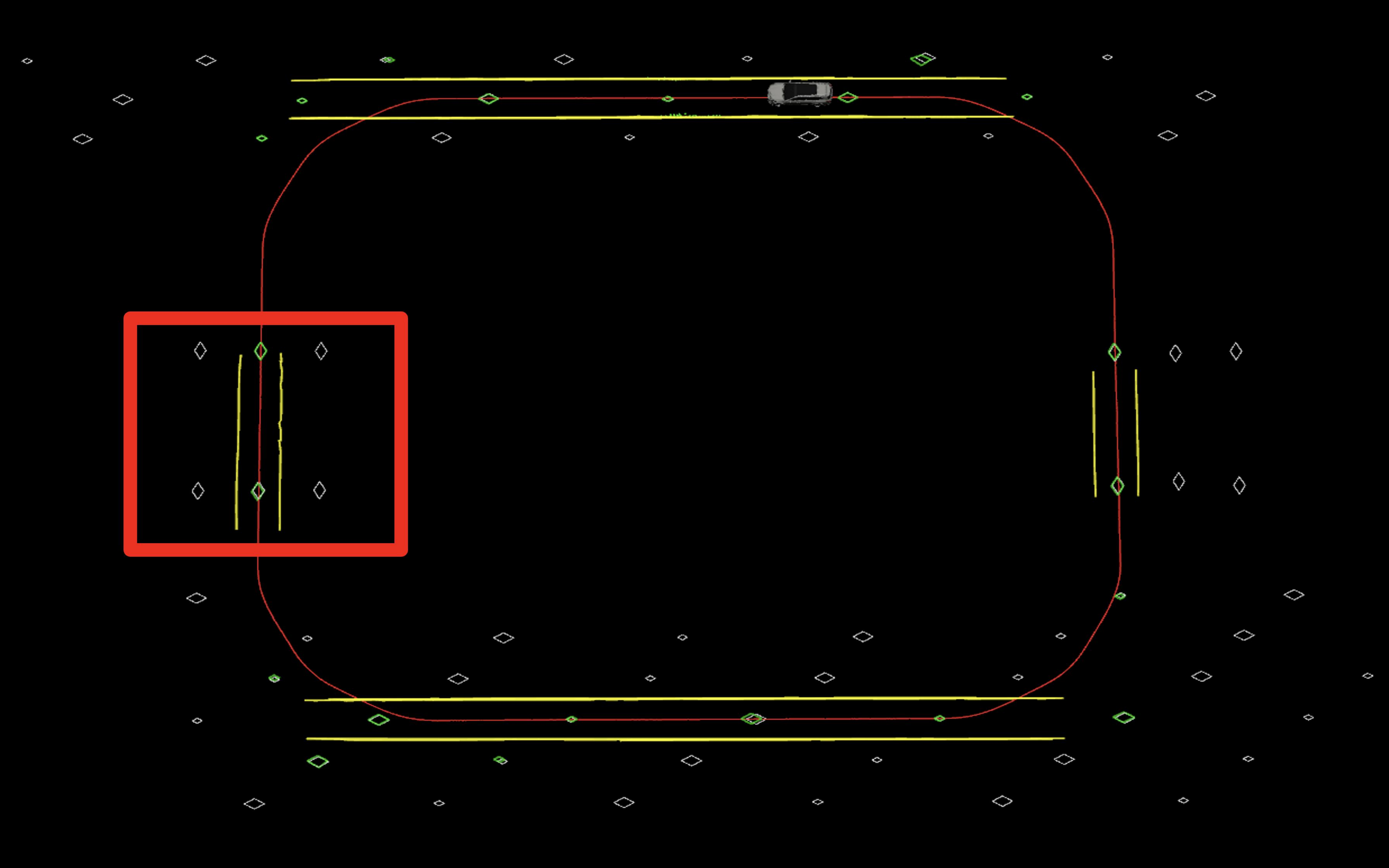}
        \label{fig: main_map}
    }
    \caption{An example of building an HD map in an automated port. Fig. \ref{fig: main_live} is an aerial photograph of the whole scene. Fig. \ref{fig: main_map} is an example of an HD map established within the coverage of the experimental trajectory. The yellow line segments are the lanes in the actual scene. The green markings are automatically generated by the proposed method. The white markings are the real markings in the scene. The positions circled in red boxes in Fig. \ref{fig: main_live} and Fig. \ref{fig: main_map} are corresponding. Please zoom in for details.}
    \vspace{-15pt}
    \label{fig: main}
\end{figure}

With the development of intelligent driving technologies, IPM has been successfully applied to many intelligent driving problems, mainly for obtaining the BEV around the vehicle. It can facilitate road and obstacle detection~\cite{oliveira2015multimodal}, free space estimation~\cite{pan2020cross}, lane keeping~\cite{yu2022accurate}, autonomous parking, local planning and even optical flow computation~\cite{mallot1991inverse}. 
However, the calibration of the IPM matrix with Total Station is also time-consuming and laborious, which brings additional costs. Therefore, proposing a more convenient calibration method without human participation will be of great value.

Given the motives mentioned above, we propose to solve the task of obtaining an accurate IPM matrix and building marking-level HD maps simultaneously. To make our method applicable to as many kinds of UGVs as possible, we seek to accomplish this task through a simple sensor setup: one or more monocular cameras. Cameras are inexpensive and easy-to-deploy sensors, which are very friendly for low-cost UGV applications, and almost all kinds of UGVs can be equipped with monocular cameras. 

Completing this task with only monocular camera(s) is quite challenging. There will be some problems if we just inversely project the target points in the image to the ground plane (naive IPM method) to build the map. First, the IPM matrix possesses a significant error if it is not accurately calibrated. Secondly, the camera's position will change slightly with the movement of the UGV, which will also bring a huge random error to the inverse projection. Hence the precision of the map constructed by naive IPM is unsatisfactory. 
Moreover, The IPM matrix has 8 degrees of freedom, which requires multiple constraints and is complex to calculate online.

We first convert the road marking corners and lane points segmented in the image into the map frame using IPM and the corresponding vehicle pose. Then we optimize the position of road markings and the IPM matrix together to make the road markings projected on the image plane as consistent with the observation as possible. 

The method can be generalized to build any road marking-level HD maps in a broader sense by changing the types of recognizable markings to the specific markings in target scenarios. For example, if an HD map is applied to open roads in urban scenes, the markings should be various common traffic markings on the road.
\subsection{Contribution}
We summarize our contribution as follows:
\begin{enumerate}
    \item We proposed a simple yet feasible pose-guided optimization framework for building a marking-level HD map with accurate markings positions using one or more monocular cameras.
    \item In the above framework, in addition to building an HD map, the IPM matrix of the used camera can also be optimized simultaneously. 
    \item We collected experimental data in two different practical application scenarios and verified the feasibility and accuracy of the method. The Root Mean Squared Error (RMSE) of the marking corners in the final marking map can be close to the centimeter level. The optimized IPM matrix can achieve the same accuracy as manual calibration.
\end{enumerate}
\begin{figure*}[htb] 
    \centering
    \includegraphics[width=0.9\textwidth]{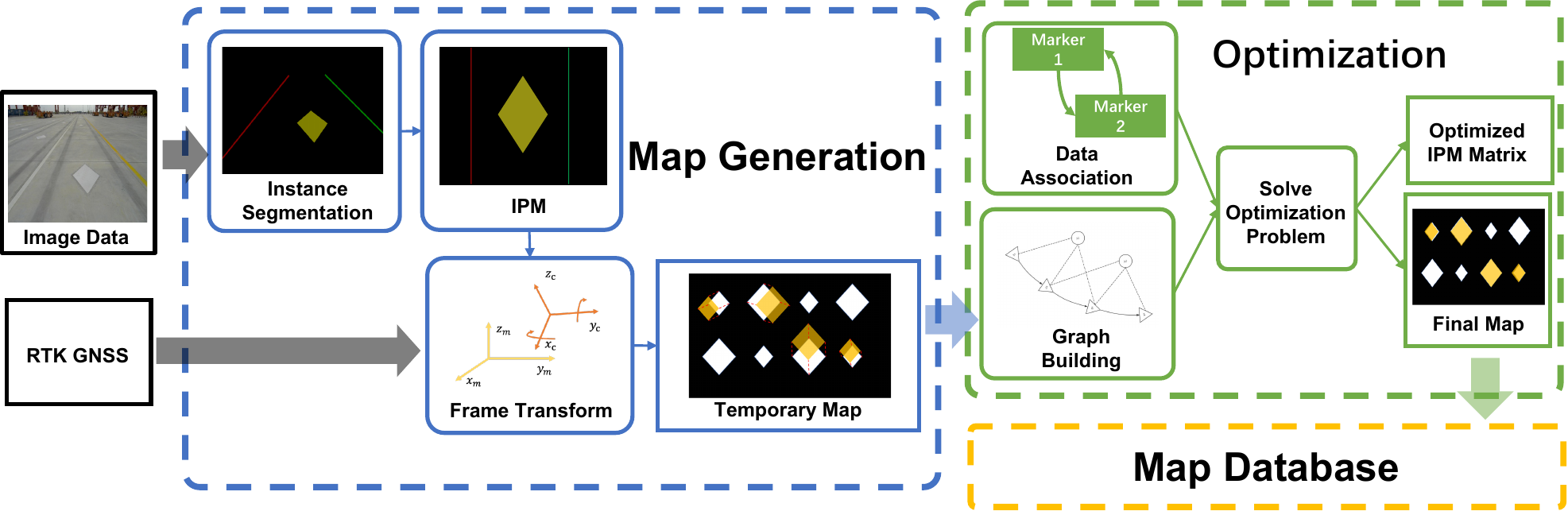}
    \caption{The inputs of our system are any number of monocular camera images and the corresponding vehicle poses obtained from RTK-GNSS. The markings extraction module recognizes the markings and lanes in the image and extracts the key points. The map generation module uses IPM to inversely project the extracted key points into the vehicle frame and then transforms them into map coordinate frame according to the vehicle pose information. Whenever a new marking is added to the map, the optimization module will conduct data association and solve the optimization problem to get the optimized map and IPM matrix.}
    \vspace{-15pt}
    \label{fig: system overview}
\end{figure*}
\section{RELATED WORK}

In recent years, there has been much research on converting image(s) perpendicular to the ground to the BEV, which is used for downstream tasks, such as local planning, road elements estimation, and semantic map construction. Yang \textit{et al.}~\cite{yang2021projecting} proposed to leverage a cycle structure and a cross-view transformer that correlates views attentively to facilitate the road scene layout estimation. Guo \textit{et al.}~\cite{guo2016low} proposed a low-cost method using normal sensor setup in contemporary cars to generate BEV images and then a lane graph of the road. The methods proposed in~\cite{lu2019monocular, mani2020monolayout, roddick2020predicting, can2022understanding, 9811901} could use only monocular camera images to generate semantic maps in the BEV.

However, the normal sensor setup limited the field of view of BEV images, so some methods used multiple images around the vehicle to reconstruct a complete surrounding map in BEV. Pan \textit{et al.}~\cite{pan2020cross} proposed VPN to parse the first-view observations from multiple angles into a BEV semantic map. Philion \textit{et al.}~\cite{philion2020lift} inferred road semantics directly in the BEV inferred from arbitrary camera rigs. Li \textit{et al.}~\cite{li2022hdmapnet} proposed HDMapNet. It can work with either or both images and point clouds. Based on semantic segmentation, vectorized maps can also be generated. Deng \textit{et al.}~\cite{deng2019restricted} proposed a method specifically designed to generate the BEV using multiple panoramic cameras. 

The results of all the above methods are limited to semantic segmentation rather than an HD map with accurate target coordinates.

HD maps include many elements, such as curbs, lane networks, road markings, semantic traffic objects, lane center lines. Recently, different methods targeting different components of the HD map are beginning to flourish. Can \textit{et al.}~\cite{can2022topology} predict the lane topology from only one on-board image. Xu \textit{et al.}~\cite{9345473}\cite{9488209}\cite{9636060}\cite{9721009} contributed a lot in road curb and boundary extraction from aerial images. Most related work focuses on establishing road, lane, and road markings maps. 

Shu \textit{et al.}~\cite{shu2020efficient} used raw crowdsourcing GPS trajectories data to build a lane-level map with both efficiency and accuracy improvement. Mi \textit{et al.}~\cite{mi2021hdmapgen} proposed a hierarchical graph generative model to generate the HD lane map in a data-driven way. 

Some researchers have tried to obtain the precise positions of the markings. Cheng \textit{et al.}~\cite{cheng2021road} extracted the sparse key points of the road markings and used Visual-Inertial Odometry (VIO) to optimize the map and the vehicle poses at the same time. 
Some methods used the 3D information from LiDAR to inversely project camera data into the point cloud to obtain HD maps. Elhousni \textit{et al.}~\cite{elhousni2020automatic} inversely projected the road surface, curb, and lane segmentation results in the images onto the point cloud directly with the calibrated transformation between the camera and LiDAR. Zhou \textit{et al.}~\cite{zhou2021automatic} inversely projected the image segmentation results onto the ground plane extracted from point cloud data. Both of them need the help of the point cloud to acquire real-world coordinates. However, in some low-cost and simple UGV applications, UGV may not be equipped with LiDAR because of the limitations of cost, space, and other factors.

Some methods similar to our idea utilized IPM to project the elements in the image into the world coordinate frame inversely. Ranganathan \textit{et al.}~\cite{ranganathan2013light} used IPM to project the front-view image to BEV inversely, then detected FAST corners within the maximally stable extremal regions (MSER). Jang \textit{et al.}~\cite{jang2018road} used the idea of graph optimization to abstract lane elements into nodes to build a lane-level HD map. Then in~\cite{jang2021lane}, they further abstracted the markings into nodes in the graph and optimized them together with the poses of the vehicle. Qin \textit{et al.}~\cite{roadmap} proposed a lightweight and highly feasible method for building HD maps. They used IPM and vehicle pose information to transform the elements segmented in the image into the map coordinate frame. Then they used graph optimization to help get better vehicle poses so that the elements' positions become more accurate accordingly. However, the IPM matrix on which these methods depend also possesses errors, so having better vehicle poses does not completely guarantee that a more accurate HD map can be established. Hence in this work, we focus on how to build a marking-level HD map with accurate markings positions and obtain an accurate IPM matrix simultaneously using one or more monocular cameras after obtaining vehicle poses.


\section{METHODOLOGY}
\begin{figure}[t] 
    \centering
    \includegraphics[width=0.3\textwidth]{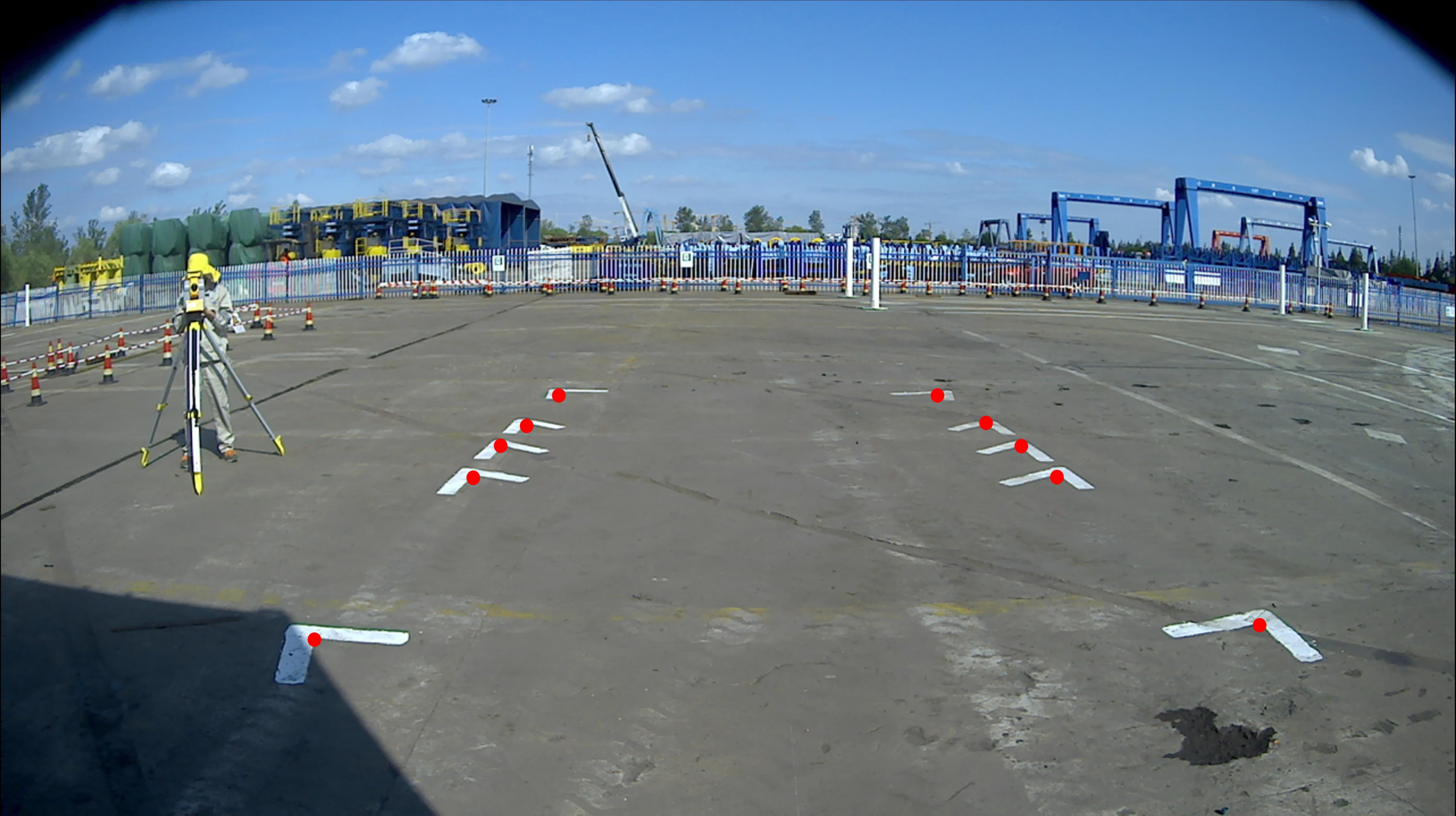}
    \caption{IPM matrix calibration example image.}
    \vspace{-15pt}
    \label{fig: cali}
\end{figure}
\subsection{Problem Statement \& System Overview}
We formulate the whole problem as a maximum likelihood estimation (MLE)~\cite{barfoot2017State} for all the measurements during the data collection run: 
\begin{equation}
    \hat{\mathcal{X}} = \argmax_{\mathcal{X}} p(\mathcal{Z}|\mathcal{X}) = \argmin_{\mathcal{X}} \sum_k f(\mathcal{X}_k,\mathbf{z}_k)
\label{eq: mle}
\end{equation}
where $\mathcal{Z}$ is the set of all measurements $\mathbf{z}_k$ for the marking corners that are independent of each other.
$\mathcal{X}$ is the set of all the variables, including the IPM matrix and the 3D coordinates of marking corners in the map frame. 
$f(\cdot)$ is the objective function. 
Assuming the measurement follows the Gaussian distribution, problem \eqref{eq: mle} can be solved as a nonlinear least-squares problem:
\begin{equation}
    \hat{\mathcal{X}} = \argmin_{\mathcal{X}} \sum_k {||\mathbf{r}(\mathcal{X}_k,\mathbf{z}_k)||}_\sigma^2,
    \label{eq: mle}
\end{equation}
where $\sigma$ is the covariance. Such a problem can be solved usually using iterative methods such as Gauss-Newton or Levenberg-Marquardt. 

In general, the inputs of our system are any number of monocular camera images and the corresponding vehicle poses. In our experimental scenarios, the vehicle poses are obtained from RTK-GNSS. The markings extraction module recognizes the markings and lanes in the image and extracts the key points. The map generation module constructs a temporary map using a naive strategy which will be elaborated in Section \ref{subsubsec: ns}. Every time a new marking is added to the map, the optimization module will optimize the location of the markings in the whole map and the IPM matrix. Refer to Fig. \ref{fig: system overview} for the framework of the whole system.
\subsection{Road Marking Detection}
In our system, we use Meta AI Research's detection and segmentation algorithms library Detectron2~\cite{wu2019detectron2} to detect the contour of the ground markings (diamonds in our experimental scenario). For the lane detection task, we use UFLD~\cite{qin2020ultra} for its remarkable speed and accuracy.
\subsection{IPM}
Following~\cite{roadmap}, we also assume that the ground is flat and all the markings are on the ground plane ($z$ equals 0). The IPM process can be defined as the following transformation~\cite{hartley2003multiple}:
\begin{equation}
    \mathbf{p}_2 = \mathbf{H} \mathbf{p}_1,
\end{equation}
expand the above formula to get:
\begin{equation}
    s
    \left( 
        \begin{array}{c}
         x  \\
         y  \\
         1
        \end{array} 
    \right)  =  
    \left( 
        \begin{array}{ccc}
         h_1 & h_2 & h_3 \\
         h_4 & h_5 & h_6 \\
         h_7 & h_8 & 1
        \end{array} 
    \right)
    \left( 
        \begin{array}{c}
         u  \\
         v \\
         1
        \end{array} 
    \right),
    \label{form: IPM}
\end{equation}
where $(x,y)$ is the coordinate of the point in the vehicle frame after the inverse projection transformation. 
\subsubsection{Pre-calibration}
\label{subsubsec: pre-cali}
We use the multiple point pairs method to calculate the above-mentioned $\mathbf{H}$ matrix. Firstly, we manually mark the target corners in the picture to obtain their pixel coordinates as in Fig. \ref{fig: cali}. On the experimental site, we invite professional surveying and mapping personnel to use Total Station to obtain the coordinates of the target corners in the vehicle frame. This way, we obtained 10 point pairs and can easily calculate the homography matrix.
\subsubsection{Naive Strategy}
\label{subsubsec: ns}
After obtaining the marking detection result, a polygon is used to fit the contour of the segmentation results in the image. Then the polygon's corners are used as the keypoints to represent the marking. The simplest way to build the map is to directly inversely project the keypoints extracted from the image into the vehicle frame by equation \eqref{form: IPM}, then transform them into the map frame according to the vehicle pose. We use the average coordinate as the final coordinate for those markings observed many times. 
Because the farther the target point is from the vehicle body, the greater the error of the IPM. Therefore, we take the region determined by the outermost calibration points, which is close to the camera center as the Region of Interest (ROI) used. Only pixels within the ROI will be inversely projected.
\subsection{Optimization}
To minimize the displacement of the inversely projected markings, instead of directly fitting the contours\cite{jang2021lane} after projecting them to the ground plane, we pursue to estimate an optimal homography matrix $\mathbf{H}$ and the position of markings' corners at the same time to minimize the projection error.
The reprojection relation from the marking corner positions in the Euclidean space to the image pixel plane follows the equation:
\begin{equation}
\lambda \begin{bmatrix}
u \\ v \\ 1
\end{bmatrix} = \mathbf{K} [\mathbf{R}_{cb} | \mathbf{t}_{cb}]  [\mathbf{R}_{wb} | \mathbf{t}_{wb}]^{-1} \begin{bmatrix}
x_w \\ y_w \\ z_w \\ 1
\end{bmatrix},
\label{eq: reprojection}
\end{equation}
where $\mathbf{K}$ is the intrinsic matrix. $[\mathbf{R}_{cb} | \mathbf{t}_{cb}]$ and $[\mathbf{R}_{wb} | \mathbf{t}_{wb}]$ are the transformation matrix of the camera extrinsic from camera frame to vehicle frame and vehicle pose, respectively.
Therefore, the homography matrix $\mathbf{H}$ which inversely project pixel coordinates from the image plane to the coordinates on the ground plane in the vehicle frame, can be written as a combination of the intrinsic and extrinsic of the camera as follows:
\begin{equation}
\begin{aligned}
    \begin{bmatrix}
x_b \\ y_b \\ 1
\end{bmatrix} &= \mathbf{H} \cdot \lambda \begin{bmatrix}
u \\ v \\ 1
\end{bmatrix}, \\
\text{with} \begin{bmatrix}
x_w \\ y_w \\ 1
\end{bmatrix} &= [\mathbf{R}_{wb}\:_{col:1,2} | \mathbf{t}_{wb}] \begin{bmatrix}
x_b \\ y_b \\ 1
\end{bmatrix}, \\
\mathbf{H} &= [\mathbf{R}_{cb}\:_{col:1,2} | \mathbf{t}_{cb}]^{-1} \mathbf{K}^{-1}.
\end{aligned}
\label{eq: h_ex_in}
\end{equation}
The third column of the rotation matrix can be omitted, leveraging the fact that the inversely projected points are on the ground plane.

\begin{figure}[t] 
    \centering
    \includegraphics[width=0.45\textwidth]{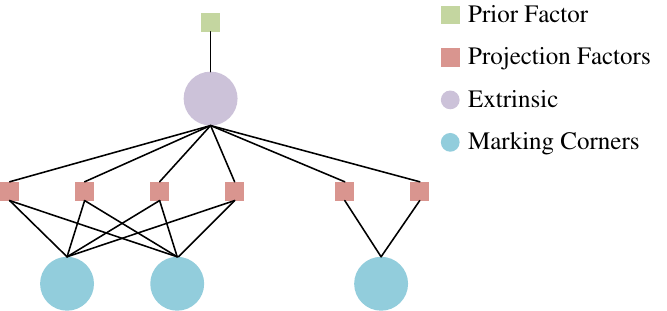}
    \caption{The illustration of the factor graph of the bundle adjustment.}
    \vspace{-15pt}
    \label{fig: factor graph}
\end{figure}

To build an HD map, we treat the problem as a bundle adjustment to estimate the homography matrix $\mathbf{H}$ and the positions of the markings simultaneously. An illustration figure of the factor graph is shown in Fig. \ref{fig: factor graph}.

As the camera's intrinsic matrix $\mathbf{K}$ is relatively easy to obtain, we assume it is well-calibrated and known. We also deem the vehicle poses $[\mathbf{R}_{wb} | \mathbf{t}_{wb}]$ acquired from RTK-GNSS accurate. The problem of optimizing $\mathbf{H}$ matrix thus can be converted to optimizing the extrinsic matrix from the camera frame to the vehicle frame. 

Since the vehicle lacks the rotation motion, which causes a degeneration of the translation part of the extrinsic matrix,
a translation prior (installation position of the camera on the vehicle), according to the installation drawings provided by the UGV manufacturer, is used to constrain the extrinsic matrix.
The initial guess of the extrinsic rotation matrix complies with the assumption that the $xy$-plane of the vehicle coordinate frame are parallel to the ground, the $z$-axis is vertical to the ground and upward, the $z$-axis of the camera coordinate frame is parallel to the ground and forward, and the $y$-axis is vertical to the ground and downward.

The initial guess of each marking corner's position is provided by the coarse IPM matrix using the naive strategy as in \ref{subsubsec: ns} at the beginning and then provided by a transformation composed of extrinsic and intrinsic as equation \eqref{eq: h_ex_in} after the first iteration.

With variable $\mathcal{\hat{X}}$ containing the estimations for camera extrinsic and marking corners, equation \eqref{eq: mle} can be elaborated as the following equation:
\begin{equation}
\begin{aligned}
\mathcal{\hat{X}}&=\{\mathbf{\hat{R}}_{cb}, \mathbf{\hat{t}}_{cb}, \mathbf{\hat{l}}_0...\mathbf{\hat{l}}_n\} \\
&=\argmin_{\mathbf{R}_{cb}, \mathbf{t}_{cb}, \mathbf{l}_0...\mathbf{l}_n} \Biggl\{ \sum_{i \in [1,4n], j \in [1,m]} f(i, j) ||\mathbf{\pi}_{ij}(\mathbf{l}_i) - \mathbf{z}_{ij}||^2_\sigma \\ &\quad + ||\mathbf{t}_{cb} - \mathbf{t}_0)||^2_\sigma \Biggl\}
\end{aligned}
\end{equation}
where $\mathbf{l}_i$ represents the 3D coordinate of the marking corner in the map frame. The total number of $\mathbf{l}_i$ is $4n$ because there are $4$ corners in each diamond marking. $f(i,j)$ equals to 1, if the marking corner $\mathbf{l}_i$ can be observed from the $j$th pose, otherwise it equals to 0. 
$\mathbf{\pi}_{ij}$ follows \eqref{eq: reprojection} projecting the marking corner $\mathbf{l}_i$ to the image pixel according to the $j$th pose. And  $\mathbf{z}_{ij}$ is the pixel measurement for $\mathbf{l}_i$ observed from $j$th pose. The prior $\mathbf{t}_0$ is used to constrain the translation part of the camera extrinsic.
\section{EXPERIMENTAL RESULTS}
\begin{figure*}[htbp] 
    \centering
    \subfigure[Rear Monocular Camera]{
        \includegraphics[width=0.31\textwidth]{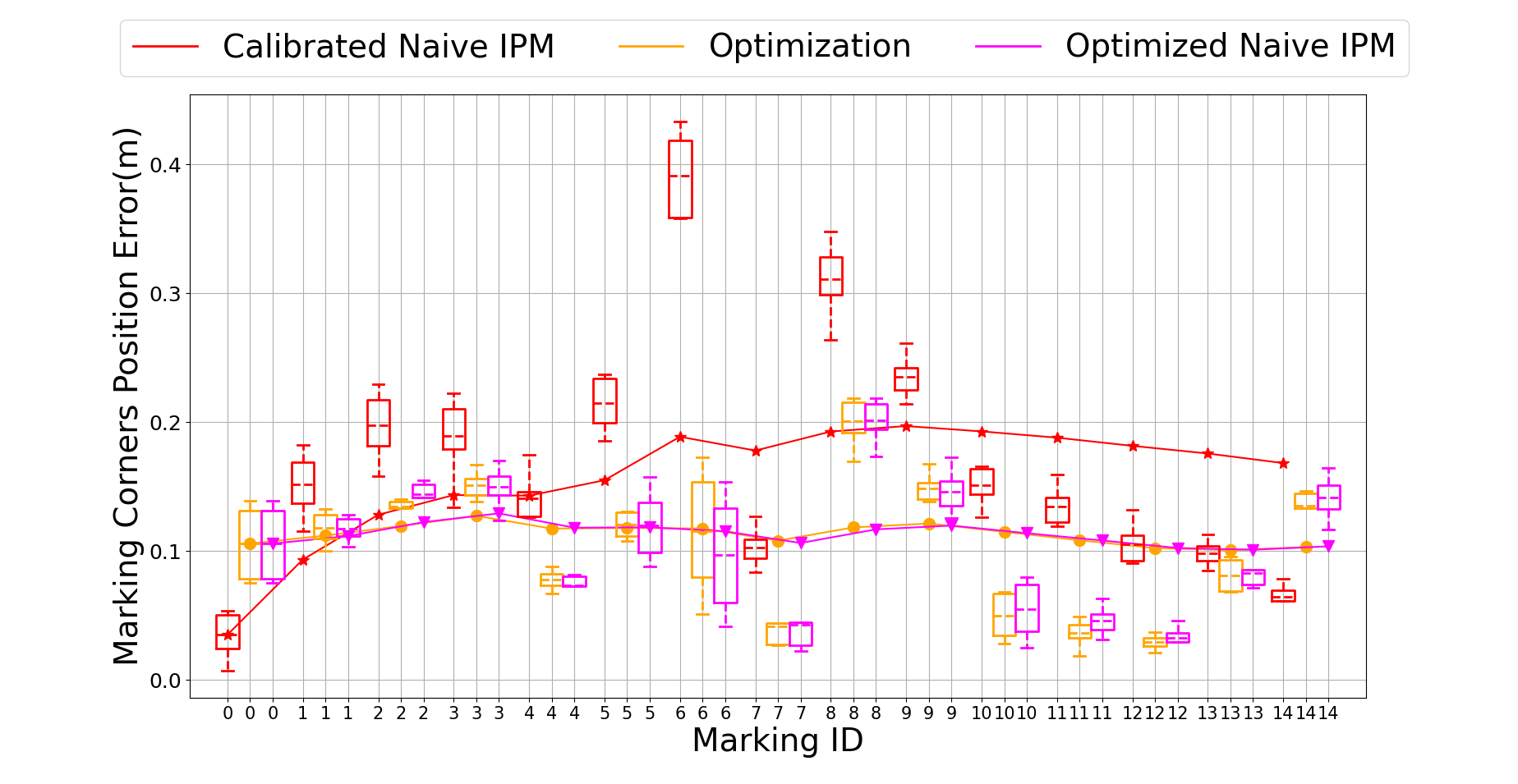}
        \label{fig: map_RMSE_error_gz_rear}
    }
    \subfigure[Front Monocular Camera]{
        \includegraphics[width=0.31\textwidth]{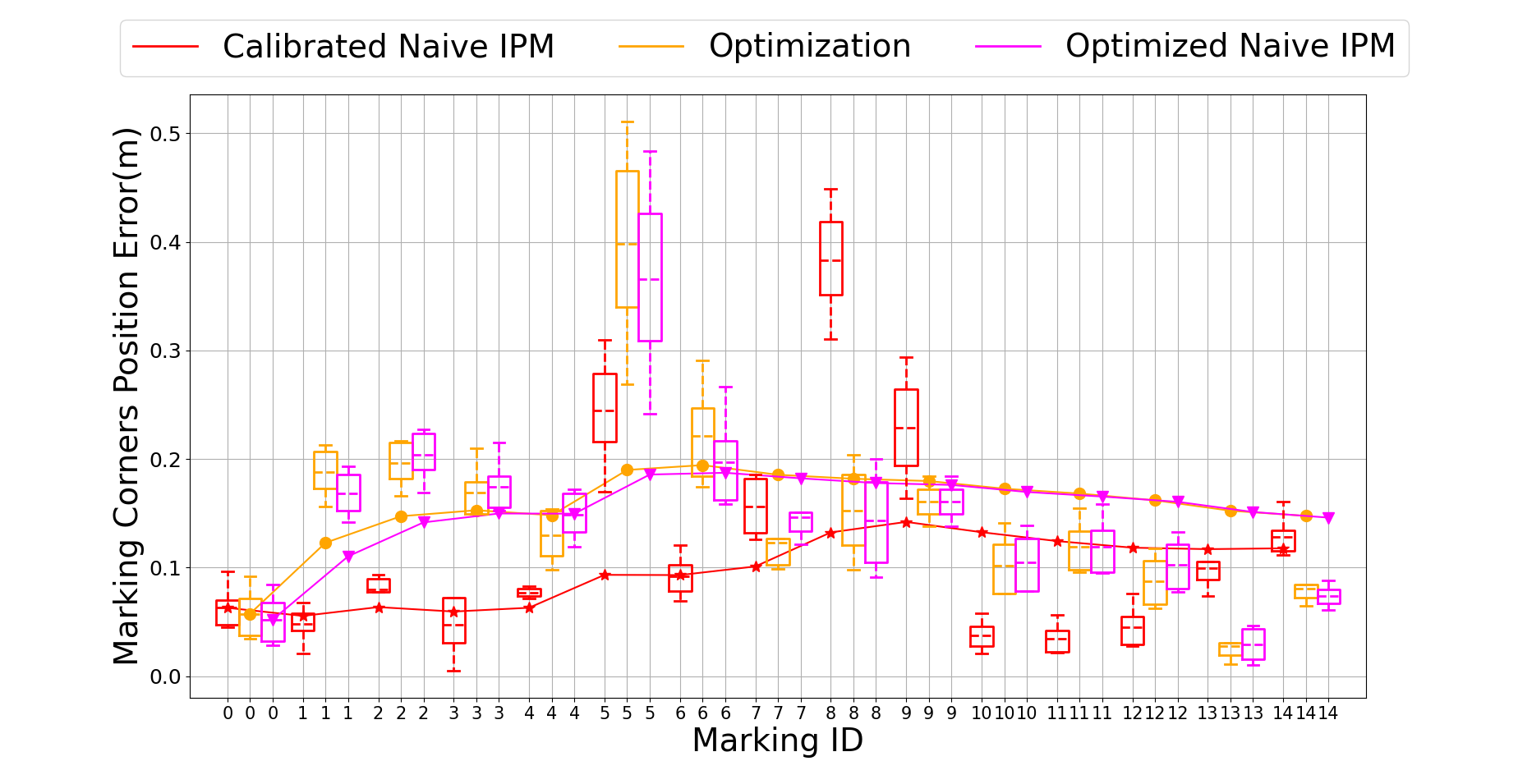}
        \label{fig: map_RMSE_error_gz_front}
    }
    \subfigure[Front and Rear Monocular Cameras]{
        \includegraphics[width=0.31\textwidth]{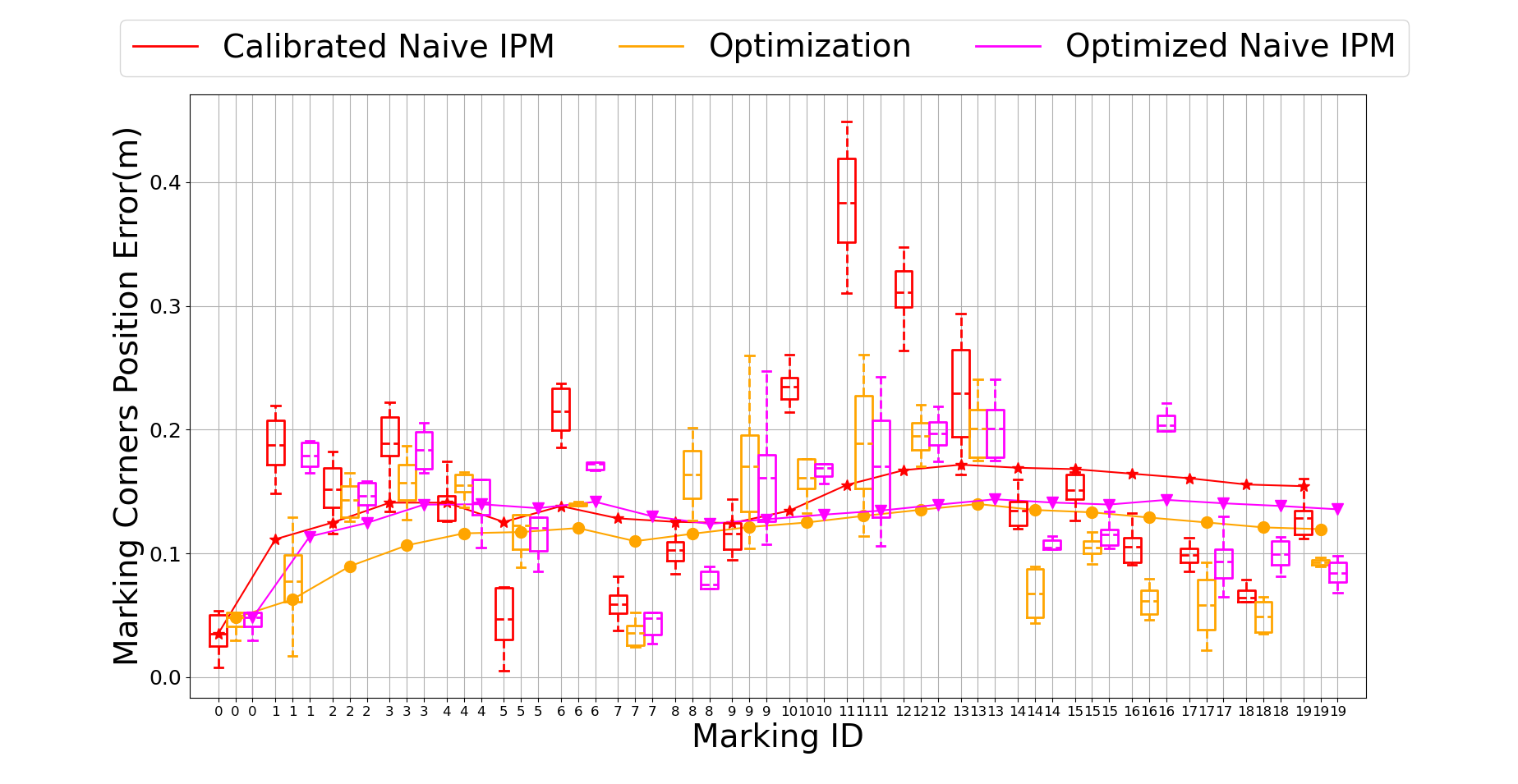}
        \label{fig: map_RMSE_error_gz_fr}
    }
    \subfigure[Rear Monocular Camera]{
        \includegraphics[width=0.31\textwidth]{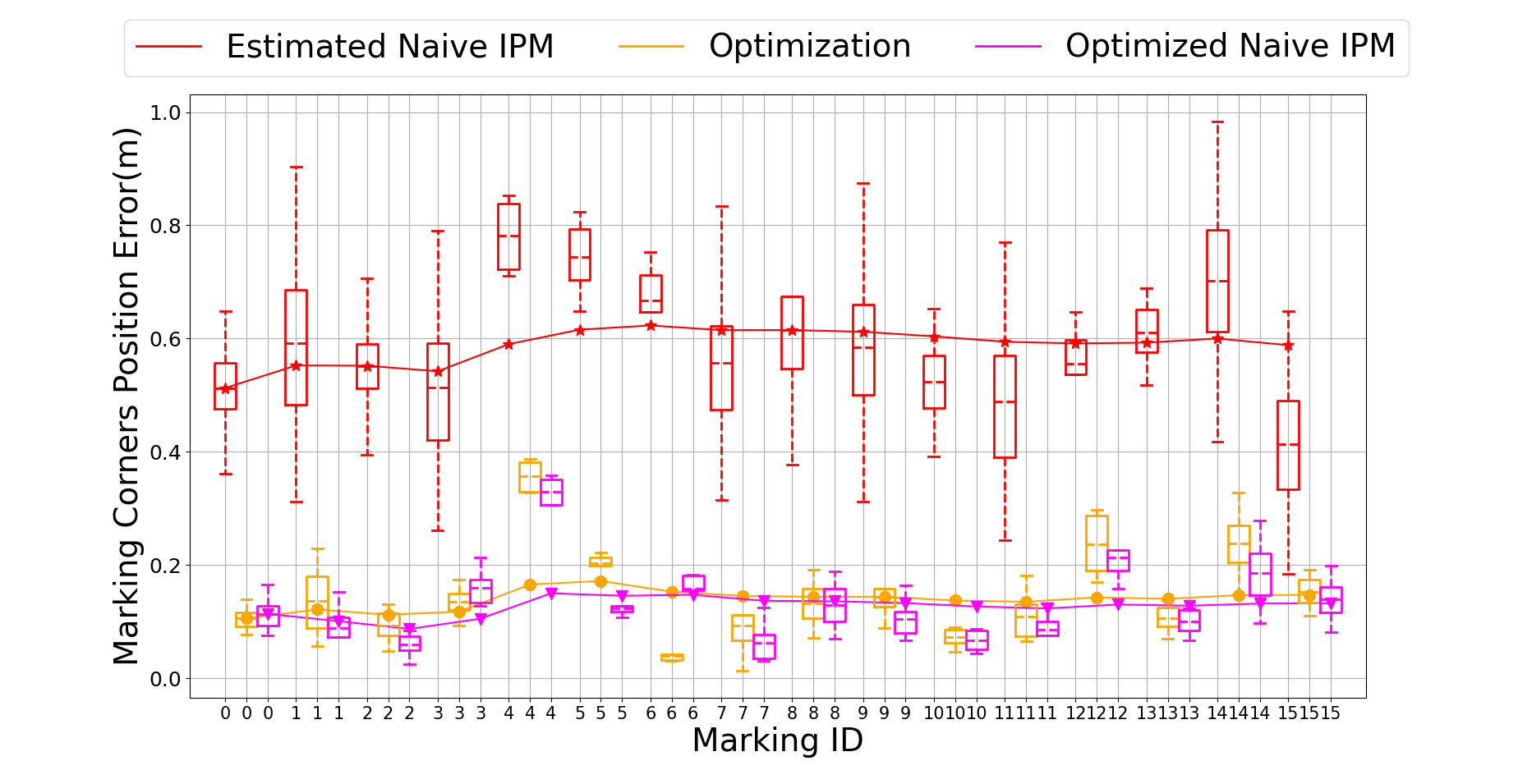}
        \label{fig: map_error_sh_rear}
    }
    \subfigure[Front Monocular Camera]{
        \includegraphics[width=0.31\textwidth]{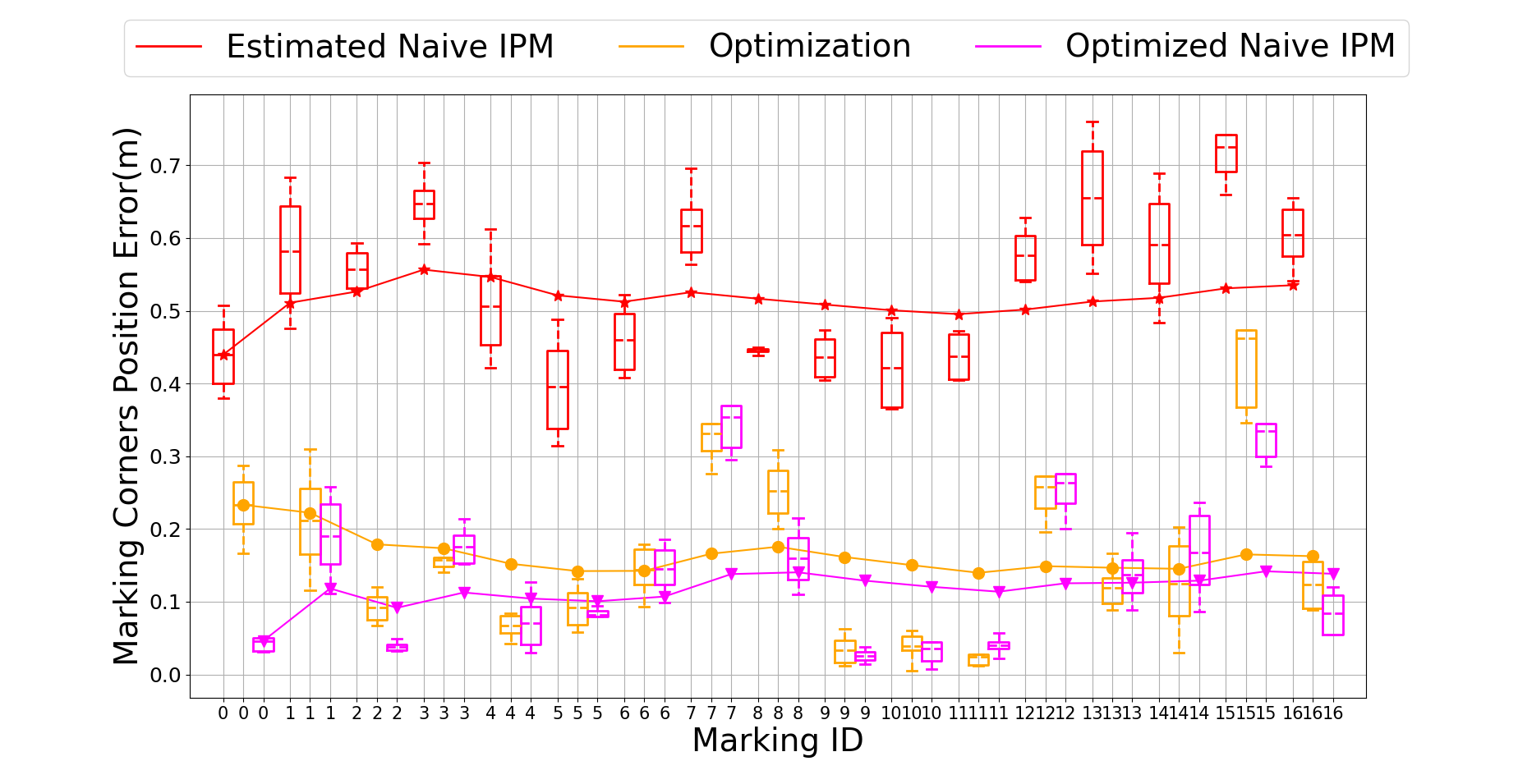}
        \label{fig: map_error_sh_front}
    }
    \subfigure[Front and Rear Monocular Cameras]{
        \includegraphics[width=0.31\textwidth]{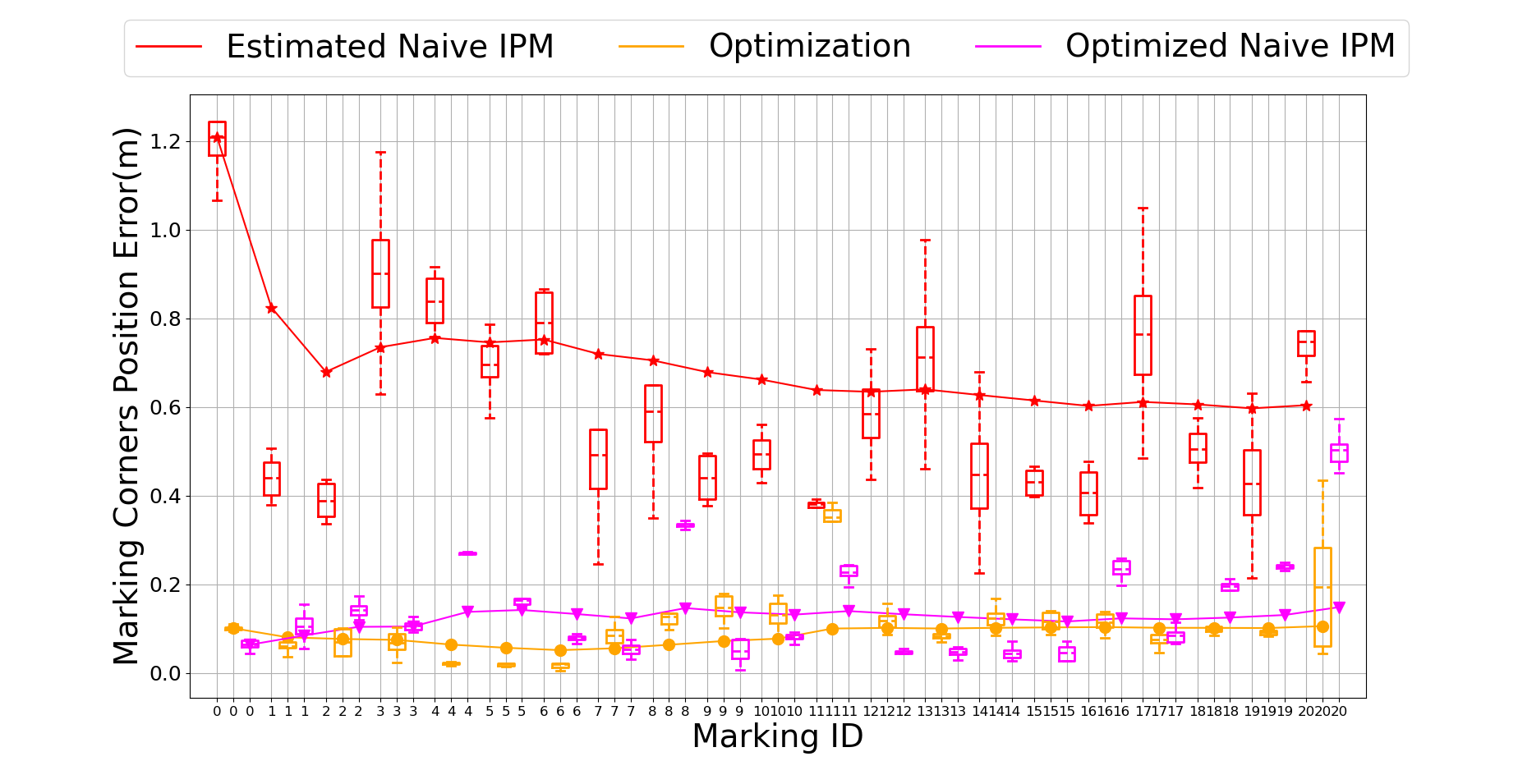}
        \label{fig: map_error_sh_fr}
    }
    \caption{Marking error comparison in two scenarios. The maps generated in the first scenario by CNI, Opt, and ONI with different monocular cameras are shown in \ref{fig: map_RMSE_error_gz_rear} \ref{fig: map_RMSE_error_gz_front} \ref{fig: map_RMSE_error_gz_fr}. The maps generated in the second scenario by ENI, Opt, and ONI with different monocular cameras are shown in \ref{fig: map_error_sh_rear} \ref{fig: map_error_sh_front} \ref{fig: map_error_sh_fr}. The Boxes represent the position error of the four corners in each marking, and the line with the corresponding color represents the change of the average corners position error of all markings with the increase of markings in the map. The position error is the Euclidean distance from the ground truth.}
    \vspace{-10pt}
    \label{fig: map error}
\end{figure*}
\subsection{Data Source}
To the best of our knowledge, the existing public HD map data sets can not provide the global coordinates ground truth for ground markings, which can be used to evaluate the accuracy of markings in the map directly. Therefore, we tested our method in two automated ports that rely on painting diamond markings on the ground to complete the visual localization of UGV. The data collected using different port UGVs includes monocular camera images and RTK-GNSS with centimeter-level accuracy. In addition, we used the Total Station to obtain the precise coordinates of all the corners of the markings on the site by manual measurement, which can provide the global coordinates ground truth for markings. 

There are hundreds of markings totally in these two fields, respectively, and dozens of markings within the coverage of the experimental data collection route. There are lanes in both areas to ensure the safe operation of UGV.
\subsection{Baselines}
Since there is still a lack of work to build an HD map with global marking coordinates based on monocular image(s) and vehicle poses, we will prove the effectiveness of the method and the accuracy of the generated HD map and IPM matrix by comparing our method with the following baselines.
\begin{enumerate}
    \item \textbf{Calibrated Naive IPM (CNI):}
        The pre-calibrated IPM matrix is used to conduct the naive strategy.
    \item \textbf{Estimated Naive IPM (ENI):}
        It is the same as the above method, but the pre-calibrated IPM matrix comes from a different UGV. It means that the IPM matrix is inaccurate because the mechanical structures of different UGVs are quite different.
    \item \textbf{Optimization (Opt):}
        Use the naive strategy to inversely project the corners of the markings into the vehicle frame, then optimize the corners' position and IPM matrix of the corresponding monocular camera.
    \item \textbf{Optimized Naive IPM (ONI):}
        Conduct the naive strategy with the IPM matrix optimized by the optimization process.
    \item \textbf{With One or More Monocular Cameras:}
        In our experiments, we use the front-mounted monocular camera, the rear-mounted monocular camera and both for comparison. When multiple cameras are used, each camera executes the above method separately.
\end{enumerate}
\subsection{Results}
\subsubsection{Generated Maps}
\begin{figure}[t] 
    \centering
    \subfigure[Estimated Naive IPM (ENI)]{
        \includegraphics[width=0.22\textwidth]{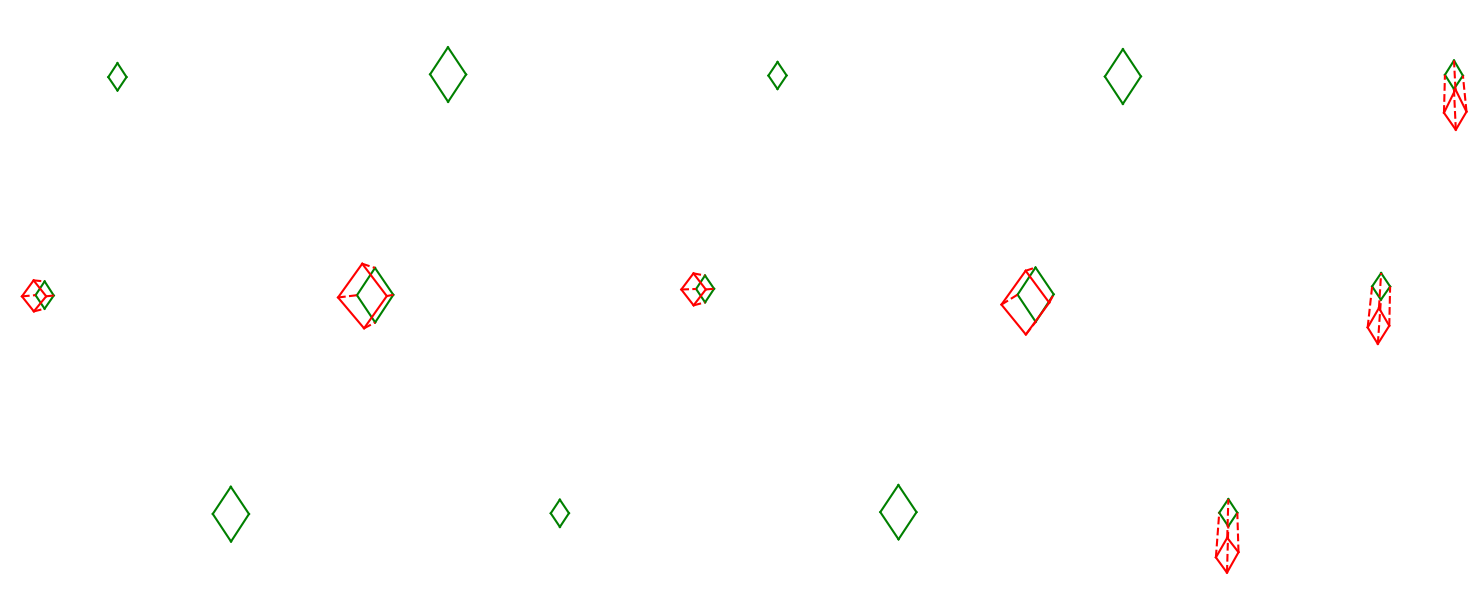}
        \label{fig: map_diff_eni}
    }
    \subfigure[Optimized Naive IPM (ONI)]{
        \includegraphics[width=0.22\textwidth]{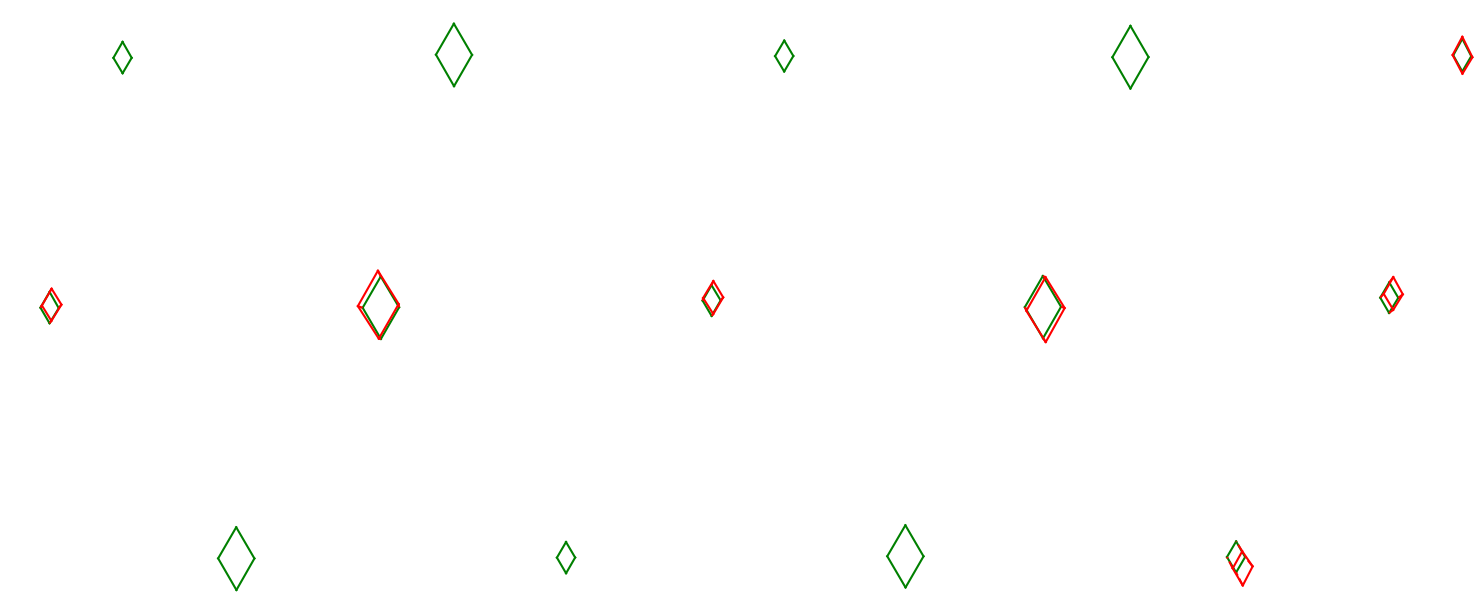}
        \label{fig: map_diff_oni}
    }
    \caption{The comparison between the maps generated by baselines and the ground truth. The green markings are the ground truth. The red markings are generated by baselines. The corner corresponding relationship is marked by red dash lines.}
    \label{fig: map diff}
\end{figure}

The marking level HD map generated by our method in our second test scenario is shown in Fig. \ref{fig: main}. 
The comparison of the partial HD map built with the inaccurate IPM matrix and the method proposed is shown in Fig. \ref{fig: map diff}. There is a significant error between the markings generated by ENI and the ground truth. The markings generated by ONI coincide with the ground truth better.
\subsubsection{Quantitative Results}
In this section, we quantitatively compare the maps generated by different baselines using the RMSE as the index. Specifically, we calculate the RMSE of the corners of all markings with ground truth to evaluate the corner accuracy of the map. 
In our first test scenario, we obtain the accurate IPM matrix for the front and rear monocular cameras using the pre-calibration method described in Section \ref{subsubsec: pre-cali}. The error comparison of the generated maps from different baselines is shown in Fig. \ref{fig: map_RMSE_error_gz_rear} \ref{fig: map_RMSE_error_gz_front} \ref{fig: map_RMSE_error_gz_fr}. 

Due to the existence of calibration error, the accuracy of the IPM matrix calibrated for different cameras is different. Under different camera setups, our optimization method can reach similar accuracy to the CNI baseline, and the RMSE of the marking corners in the map is close to the centimeter level. Using the ONI method can achieve similar accuracy to the CNI shows that our method can obtain an IPM matrix with similar accuracy to the manual calibration while building the map. Thus, our method can replace the tedious process of manually calibrating the IPM matrix.

We used a different UGV to collect data in our second test scenario but still use the same IPM matrix. Due to the different mechanical structures of different vehicles, this IPM matrix is totally inaccurate. Under such conditions, the error comparison between generated maps from different methods is shown in Fig. \ref{fig: map_error_sh_rear} \ref{fig: map_error_sh_front} \ref{fig: map_error_sh_fr}. 

It can be seen from the results that the accuracy of ENI is much lower because of the inaccurate IPM matrix. However, our optimization method can still build a map with an accuracy close to the centimeter level. ONI can achieve the same level of accuracy as CNI shows that even if we do not provide an accurate IPM matrix beforehand, the accuracy of the generated map and optimized IPM matrix are not affected. This further reduces the difficulty of large-scale deployment of our method because it is not necessary to calibrate the IPM matrix for each vehicle anymore. Instead, we can use a coarse one as a prior and optimize it during the mapping process for each vehicle.

The accuracy of using multiple monocular cameras is similar to using a single one. But using multiple monocular cameras will observe more markings, improving HD map construction efficiency.

\section{CONCLUSIONS \& FUTURE WORK}
This paper proposes a pose-guided optimization framework for building a marking-level HD map with accurate markings positions using one or more monocular cameras, in which the homography matrix for IPM and the map are optimized at the same time. We evaluate the map accuracy in two test scenarios. The marking corner coordinates accuracy can be close to centimeter level, and the IPM matrix accuracy is similar to manually calibrated. We highlight that it is a feasible and easy-to-extend solution to building HD maps at a low cost. In the future, we will extend the marking types to a wider range and generalize the method further.










\bibliographystyle{IEEEtran}
\bibliography{egbib}

\end{document}